\documentclass[10pt,twocolumn,letterpaper]{article}

\usepackage{cvpr}
\usepackage{times}
\usepackage{epsfig}
\usepackage{graphicx}
\usepackage{amsmath}
\usepackage{amssymb}

\usepackage{multirow}
\usepackage{epstopdf}
\usepackage{amsfonts}
\usepackage{algorithm}
\usepackage{algorithmic}
\usepackage{diagbox}

\usepackage{subfigure}
\def\ds3c{\textbf{D$\text{S}^3$C}}
\def\s3c{$\text{S}^3$C}

\def\e{\textbf{e}}

\def\h{\textbf{h}}
\def\q{\textbf{q}}
\def\w{\textbf{w}}
\def\x{\textbf{x}}
\def\y{\textbf{y}}
\def\z{\textbf{z}}
\def\0{\textbf{0}}
\def\1{\textbf{1}}
\def\ie{i.e.}
\def\eg{e.g.}
\def\st{\textrm{s.t.}}
\def\and{\textrm{and}}
\def\rank{\textrm{rank}}

\def\diag{\textrm{diag}}

\def\Q{\mathcal{Q}}
\newcommand{\RR}{I\!\!R} %real numbers

\newcommand{\myparagraph}[1]{\smallskip\noindent\textbf{#1.}}

\def\ie{i.e.}
\def\eg{e.g.}

\cvprfinalcopy % *** Uncomment this line for the final submission

\begin{document}

%%%%%%%%% TITLE
\title{Self-Supervised Convolutional Subspace Clustering Network}

%\author{Junjian Zhang\\
%% omit the following lines up until the closing ``}''.
%% Additional authors and addresses can be added with ``\and'',
%% just like the second author.
%% To save space, use either the email address or home page, not both
%\and

\author{Junjian Zhang$^\dag$, Chun-Guang Li$^\dag$,~Chong You$^\ddag$,~Xianbiao Qi$^\sharp$,\\
~Honggang Zhang$^\dag$,~Jun Guo$^\dag$,~and~Zhouchen Lin$^\S$\\
%$^1$ School of Info. \& Commu. Engineering, Beijing Univ. of Posts and Telecommunications, P. R. China \\
%$^1$ School of Info. \& Commu. Engineering, Beijing University of Posts and Telecommunications \\
$^\dag$ SICE, Beijing University of Posts and Telecommunications \\
$^\ddag$ EECS, University of California, Berkeley \ \ \ \ $^\sharp$ Shenzhen Research Institute of Big Data \\ %Cooperative Medianet Innovation Center, Shanghai Jiaotong University \\
$^\S$ Key Laboratory of Machine Perception (MOE), School of EECS, Peking University \\
%$^4$ School of EECS, Peking University \\
%{\tt\small \{zhjj, lichunguang, zhhg, guojun\}@bupt.edu.cn; zlin@pku.edu.cn}
%Institution1 address
}

\maketitle

%%%%%%%%% ABSTRACT
\begin{abstract}

Subspace clustering methods based on data self-expression have become very popular for learning from data that lie in a union of low-dimensional linear subspaces. However, the applicability of subspace clustering has been limited because practical visual data in raw form do not necessarily lie in such linear subspaces. On the other hand, while Convolutional Neural Network (ConvNet) has been demonstrated to be a powerful tool for extracting discriminative features from visual data, training such a ConvNet usually requires a large amount of labeled data, which are unavailable in subspace clustering applications. To achieve simultaneous feature learning and subspace clustering, we propose an end-to-end trainable framework, called Self-Supervised Convolutional Subspace Clustering Network (S$^2$ConvSCN), that combines a ConvNet module (for feature learning), a self-expression module (for subspace clustering) and a spectral clustering module (for self-supervision) into a joint optimization framework. Particularly, we introduce a dual self-supervision that exploits the output of spectral clustering to supervise the training of the feature learning module (via a classification loss) and the self-expression module (via a spectral clustering loss). Our experiments on four benchmark datasets show the effectiveness of the dual self-supervision and demonstrate superior performance of our proposed approach.

\end{abstract}

%\linenumbers

\section{Introduction}
\label{sec:intro}

In many real-world applications such as image and video processing, we need to deal with a large amount of high-dimensional data.
Such data can often be well approximated by a union of multiple low-dimensional subspaces, where each subspace corresponds to a class or a category. For example, the frontal facial images of a subject taken under varying lighting conditions approximately span a linear subspace of dimension up to nine \cite{Ho:CVPR03}; the trajectories of feature points related to a rigidly moving object in a video sequence span an affine subspace of dimension up to three \cite{CT:IJCV92}; the set of handwritten digit images of a single digit also approximately span a low-dimensional subspace \cite{Hastie:StatSci98}. In such cases, it is important to segment the data into multiple groups where each group contains data points from the same subspace. %, which relates to the meaningful structures (or patterns) in datasets.
This problem is known as \textit{subspace clustering} \cite{ReVidal:ISM11,Vidal:Springer16}, which we formally define as follows.

{\textbf{Problem} (Subspace Clustering).} {\em Let $X \in \RR^{D \times N}$ be a real-valued matrix whose columns are drawn from a union of $n$ subspaces of $\RR^D$, $\bigcup_{i=1}^n \{S_i\}$, of dimensions $d_i \ll \min \{D, N\}$, for $i=1,\dots,n$. The goal of subspace clustering is to segment the columns of $X$ into their corresponding subspaces.}

In the past decade, subspace clustering has become an important topic in unsupervised learning and many subspace clustering algorithms have been developed \cite{Elham:CVPR09, Liu:ICML10, Favaro:CVPR11, Lu:ECCV12, Liu:TPAMI13, Lu:ICCV13-TraceLasso, Feng:CVPR14, Li:CVPR15, Peng:TCYB16, Zhang:VCIP16, You:CVPR16-EnSC, Li:TIP17}. % , Lu:TPAMI18
These methods have been successfully applied to various applications such as motion segmentation \cite{Vidal:PAMI05, Rao:TPAMI10}, face image clustering~\cite{Elham:TPAMI13}, genes expression microarray clustering \cite{McWilliams:DMKD14, Li:ICPR18} and so on.

Despite the great success in the recent development of subspace clustering, its applicability to real applications is very limited because practical data do not necessarily conform with the linear subspace model.
In face image clustering, for example, practical face images are often not aligned and often contain variations in pose and expression of the subject.
Subspace clustering cannot handle such cases as images corresponding to the same face no longer lie in linear subspaces.
While there are recently developed techniques for joint image alignment and subspace clustering \cite{Li:PR16}, such a parameterized model is incapable of handling a broader range of data variations such as deformation, translation and so on.
It is also possible to use manually designed invariance features such as SIFT~\cite{Lowe:IJCV04}, HOG \cite{Dalal:CVPR05} and PRICoLBP~\cite{Qi:TPAMI14} of the images before performing subspace clustering, \eg, in \cite{Peng:IJCAI16, Peng:arxiv17}.
However, there has been neither theoretical nor practical evidence to show that such features follow the linear subspace model.

Recently, Convolutional Neural Networks (ConvNets) have demonstrated superior ability in learning useful image representations in a wide range of tasks such as face/object classification and detection \cite{Krizhevsky:NIPS2012,Parkhi:BMVC15}.
In particular, it is shown in \cite{Lezama:CVPR18} that when applied to images of different classes, ConvNets are able to learn features that lie in a union of linear subspaces.
The challenge for training such a ConvNet, however, is that it requires a large number of labeled training images which is often unavailable in practical applications.
	
In order to train ConvNet for feature learning without labeled data, many methods have been recently proposed by exploiting the self-expression of data in a union of subspaces \cite{Peng:IJCAI16, Ji:NIPS17, Peng:arxiv17,Zhou:CVPR18}.
Specifically, these methods supervise the training of ConvNet by inducing the learned features to be such that each feature vector can be expressed as a linear combination of the other feature vectors.
However, it is difficult to learn good feature representations in such an approach due to the lack of effective supervision.

% OLE: Orthogonal Low-rank Embedding: A Plug and Play Geometric Loss for Deep Learning

\myparagraph{Paper contribution}
In this paper, we develop an end-to-end trainable framework for simultaneous feature learning and subspace clustering, called Self-Supervised Convolutional Subspace Clustering Network (S$^2$ConvSCN).
In this framework, we use the current clustering results to self-supervise the training of feature learning and self-expression modules, which is able to significantly improve the subspace clustering performance.
%, in which a suitable feature learning module, a self-expression module, and a spectral clustering module are integrated into a joint optimization problem.
%In particular, in order to improve the training of the feature learning and self-expression modules,
In particular, we introduce the following two self-supervision modules:
\begin{enumerate}
\item We introduce a spectral clustering module which uses the current clustering results to supervise the learning of the self-expression coefficients. This is achieved by inducing the affinity generated from the self-expression to form a segmentation of the data that aligns with the current class labels generated from clustering.
    % cluster labels.
\item We introduce a classification module which uses the current clustering results to supervise the training of feature learning.
This is achieved by minimizing %the cross entropy
the classification loss between the output of a classifier trained on top of the feature learning module and the current class labels generated from clustering.
    % cluster labels.
    %The output of spectral clustering provides pseudo class labels which are used to supervise the feature learning through classification loss. We call this a \emph{dual self-supervision}.
\end{enumerate}
We propose a training framework where the feature representation, the data self-expression and the data segmentation are jointly learned and alternately refined in the learning procedure.
Conceptually, the initial clustering results do not align exactly with the true data segmentation, therefore the initial self-supervision incurs errors to the training.
Nonetheless, the feature learning is still expected to benefit from such self-supervision as there are data with correct labels that produce useful information. An improved feature representation subsequently helps to learn a better self-expression and consequently produce a better data segmentation (\ie, with less wrong labels).
Our experiments on four benchmark datasets demonstrate superior performance of the proposed approach. %and yield the state-of-the-art results.

\myparagraph{Paper Outline} The remainder of this paper is organized as follows. Section~\ref{sec:related-work} reviews the relevant work. Section~\ref{sec:ConvSSCNet} presents our proposal---the components in S$^2$ConvSCN, the used cost functions, and the training strategy. Section~\ref{sec:experiments} shows experimental results with discussions, and Section~\ref{sec:conclusion} concludes the paper.

\section{Related Work}
\label{sec:related-work}

In this section, we review the relevant prior work in subspace clustering. For clarity, we group them into two categories: a) subspace clustering in original space; and b) subspace clustering in feature space.

\begin{figure*}[t]
\vskip -5pt
\begin{center}
\centerline{\includegraphics[width=0.85\textwidth]{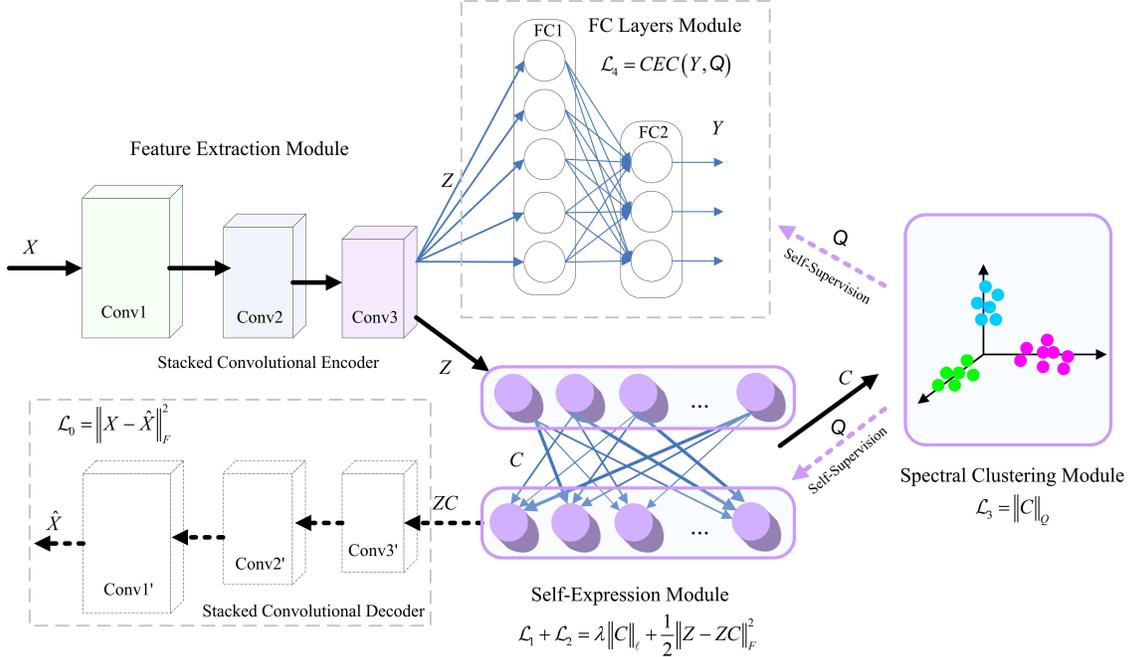}} % ConvSCNet7.eps
%\vskip - 0.15in
\caption{Architecture of the proposed Self-Supervised Convolutional Subspace Clustering Network (S$^2$ConvSCN). It consists of mainly five modules: a) stacked convolutional encoder module, which is used to extract convolutional features; b) stacked convolutional decoder module, which is used with the encoder module to initialize the convolutional module; c) self-expression module, which is used to learn the self-expressive coefficient matrix and also takes the self-supervision information from the result of spectral clustering to refine the self-expressive coefficients matrix; d) FC-layers based self-supervision module, which builds a self-supervision path back to the stacked convolutional encoder module; e) spectral clustering module, which provides self-supervision information to guide the self-expressive model and FC-layers module. The modules with solid line box are the backbone components; whereas the modules in dashed box are the auxiliary components to facilitate the training of the whole network.}
\label{fig:ConvSSCNet}
\end{center}
%\vskip -0.2in
\vskip -25pt
%\vskip -15pt
\end{figure*}

\subsection{Subspace Clustering in Original Space}
In the past years, subspace clustering has received a lot of attention and many methods have been developed. Among them, methods based on spectral clustering are the most popular, \eg, \cite{Elham:CVPR09, Liu:ICML10, Favaro:CVPR11, Lu:ECCV12, Elham:TPAMI13, Liu:TPAMI13, Lu:ICCV13-TraceLasso, Feng:CVPR14, Li:CVPR15, Peng:TCYB16, You:CVPR16-EnSC, Zhang:VCIP16, Li:TIP17, You:ECCV18}. These methods divide the task of subspace clustering into two subproblems. The %To be more specific,
first subproblem is to learn a data affinity matrix %is learned
from the original data, and the second subproblem is to apply spectral clustering on the affinity matrix to find the segmentation of the data. The two subproblems are solved successively in one-pass~\cite{Elham:CVPR09, Liu:ICML10, Favaro:CVPR11, Lu:ECCV12, Lu:ICCV13-TraceLasso, You:CVPR16-EnSC} or solved %successively but
alternately in multi-pass~\cite{Feng:CVPR14, Li:CVPR15, Guo:IJCAI15, Zhang:VCIP16, Li:TIP17}. % , Chen:PR18  , Lu:TPAMI18-BDR, Chen:PR18

Finding an informative affinity matrix is the most crucial step. Typical methods to find an informative affinity matrix are based on the self-expression property of data \cite{Elham:CVPR09, Vidal:Springer16}, which states that a data point in a union of subspaces can be expressed as a linear combination\footnote{If data points lie in a union of affine subspaces \cite{Li:JSTSP18}, then the {\em linear} combination will be modified to {\em affine} combination.} of other data points, \ie, $\x_j = \sum_{i \neq j} c_{ij} \x_i + \e_j$, where $\e_j$ is used to model the noise or corruption in data. It is expected that the linear combination of data point $\x_j$ uses the data points that belong to the same subspace as $\x_j$. To achieve this objective, different types of regularization terms on the linear combination coefficients are used. For example, in \cite{Elham:CVPR09} the $\ell_1$ norm is used to find sparse linear combination; in \cite{Liu:ICML10} the nuclear norm of the coefficients matrix is used to find low-rank representation; in \cite{Wang:NIPS13-LRR+SSC, You:CVPR16-EnSC} the mixture of the $\ell_1$ norm and the $\ell_2$ norm or the nuclear norm is used to balance the sparsity and the denseness of the linear combination coefficients; and in \cite{Xin:TSP18} a data-dependent sparsity-inducing regularizer is used to find sparse linear combination. On the other hand, different ways to model the noise or corruptions in data have also been investigated, \eg, the vector $\ell_1$ norm %and $\ell_2$ norm
is used in \cite{Elham:CVPR09}, the $\ell_{2,1}$ norm is adopted in \cite{Liu:ICML10}, and the correntropy term is used in \cite{He:TNNLS16}.
%, the Gaussian mixture model used in \cite{Li:CVPR15MoG}

\subsection{Subspace Clustering in Feature Space}

For subspace clustering in feature space, we further divide the %ways to form the feature space
existing methods into two types. The first type uses \emph{latent feature space}, which is induced via a Mercer kernel, \eg, \cite{Patel:ICIP14, Patel:JSTSP15, Ngu:Neuc15, Xiao:TNNLS16}, or constructed via matrix decomposition, \eg, \cite{Liu:ICCV11}, \cite{Patel:ICCV13}. % (Kernel SSC; Kernel LRR; ...) or constructed via matrix decomposition (Patel, Latent SSC; Hongyang et al. Latent LRR; Guangcan Liu Latent LRR; Patel:ECCV12), and
The second type use \emph{explicit feature space}, which is designed by manual feature extraction, \eg, \cite{Peng:IJCAI16}, or is learned from data, \eg, \cite{Ji:NIPS17, Zhou:CVPR18}.

\myparagraph{Latent Feature Space} Many recent works have employed the kernel trick to map the original data into a high-dimensional latent feature space, in which subspace clustering is performed, \eg, \cite{Patel:ICIP14, Patel:JSTSP15, Ngu:Neuc15, Xiao:TNNLS16}. For example, predefined polynomial and Gaussian kernels are used in the kernel sparse subspace clustering method \cite{Patel:ICIP14, Patel:JSTSP15} and the kernel low-rank representation method \cite{Ngu:Neuc15, Xiao:TNNLS16, Ji:AdaLRK-arXiv17}. Unfortunately, it is {\em not} guaranteed that the data in the latent feature space induced with such predefined kernels lie in low-dimensional subspaces.\footnote{In \cite{Ji:AdaLRK-arXiv17}, while the data matrix in the latent feature space is encouraged to be low-rank, it is not necessary that the data in feature space are encouraged to align with a union of linear subspaces.}

%(We should also discuss the work of Ji Pan: Adaptive Low-Rank Kernel Subspace Clustering)

On the other hand, the latent feature space has also been constructed via matrix decomposition, \eg, \cite{Liu:ICCV11}, \cite{Patel:ICCV13}. In \cite{Liu:ICCV11}, a linear transform matrix and a low-rank representation are computed simultaneously; in \cite{Patel:ICCV13}, a linear transform and a sparse representation are optimized jointly. %However, %there is no justification that
% why and how the learned linear transform could bring performance improvement are still unclear.
However, the representation power of the learned linear transform is still limited.

\myparagraph{Explicit Feature Space} Deep learning has gained a lot of research interests due to its powerful ability to learn hierarchical features in an end-to-end trainable way~\cite{Hinton:ISPM12, Krizhevsky:NIPS2012}. Recently, there are a few works that use techniques in deep learning for %attempts to borrow deep learning to conduct
feature extraction in subspace clustering. For example, in \cite{Peng:IJCAI16,Peng:arxiv17}, a fully connected deep auto-encoder network with hand-crafted features (\eg, SIFT or HOG features) combined with a sparse self-expression model is developed; in \cite{Ji:NIPS17}, a stacked convolutional auto-encoder network with a plus-in self-expression model is proposed. While promising clustering accuracy has been reported, these methods are still suboptimal because \emph{neither} the potentially useful supervision information from the clustering result has been taken into the feature learning %extraction
step \emph{nor} a joint optimization framework for fully combining feature learning %extraction
and subspace clustering has been developed. More recently, in \cite{Zhou:CVPR18}, a deep adversarial network with a subspace-specific generator and a subspace-specific discriminator is adopted in the framework of \cite{Ji:NIPS17} for subspace clustering. However, the discriminator need to use the dimension of each subspace, which is usually unknown.

% Dizaji:ICCV17 deep clustering

In this paper, we attempt to develop a joint optimization framework for combining feature learning %extraction
and subspace clustering, such that the useful %valuable
self-supervision information from subspace clustering result could be used to guide the feature learning %extraction
and to refine the self-expression model. Inspired by the success of Convolutional Neural Networks in recent years for classification tasks on images and videos datasets~\cite{Krizhevsky:NIPS2012} and the recent work \cite{Ji:NIPS17}, we integrate the convolutional feature extraction module into subspace clustering to form an end-to-end trainable joint optimization framework, called Self-Supervised Convolutional Subspace Clustering Network (S$^2$ConvSCN). In S$^2$ConvSCN, both the stacked convolutional layers based feature extraction and the self-expression based affinity learning are effectively self-supervised by exploiting the feedback from spectral clustering.
% We design a two-stage procedure based approach to train the proposed framework and conduct extensive experiments on benchmark datasets. Experimental results demonstrate superior performance of our proposal. % and is easier to train.

\section{Our Proposal: Self-Supervised Convolutional Subspace Clustering Network} % (ConvSSCNet)}
\label{sec:ConvSSCNet}

In this section, we present our S$^2$ConvSCN for joint feature learning and subspace clustering.
We start with introducing our network formulation (see Fig.~\ref{fig:ConvSSCNet}), then introduce the self-supervision modules.
Finally, we present an effective %algorithm
procedure for training the proposed network.

\subsection{Network Formulation}

As aforementioned, our network is composed of a feature extraction module, a self-expression module and self-supervision modules for training the former two modules.

\myparagraph{Feature Extraction Module} A basic component of our proposed S$^2$ConvSCN is the feature extraction module, which is used to extract features from raw data that are amenable to subspace clustering.
To extract localized features while preserving spatial locality, we adopt the convolutional neural network which is comprised of multiple convolutional layers.
We denote the input to the network as $\h^{(0)} = \x$ where $\x$ is the image.
A convolutional layer $\ell$ contains a set of filters $\w_i^{(\ell)}$ and the associated biases $\mathbf{b}_i^{(\ell)}$, $i = 1, \cdots, m^{(\ell)}$, and produces $m^{(\ell)}$ feature maps from the output of the previous layer. The feature maps $\{ \h_i^{(L)} \}_{i=1,\cdots,m^{(L)}}$ in the top layer $L$ of the network are then used to form a representation of the input data $\x$. Specifically, the $m^{(L)}$ feature maps $\{\h_i^{(L)}\}_{i=1}^{m^{(L)}}$ are vectorized and concatenated to form a representation vector $\z$, i.e.,
\begin{align}
\begin{split}
\z = \begin{bmatrix}  \h_1^{(L)}(:),\cdots, \h_{m^{(L)}}^{(L)}(:) \end{bmatrix}^\top,
\end{split}
\label{eq:latent-representation-z}
\end{align}
where $ \h_1^{(L)}(:),\cdots, \h_{m^{(L)}}^{(L)}(:)$ are row vectors denoting the vectorization of the feature maps $\h_1^{(L)},\cdots,\h_{m^{(L)}}^{(L)}$.
These vectors are horizontally concatenated and then transposed to form the vector $\z$.

To ensure that the learned representation $\z$ contains meaningful information from the input data $\x$, %such representation is
the feature maps $\h_1^{(L)},\cdots,\h_{m^{(L)}}^{(L)}$ are fed into a decoder network to reconstruct an image $\hat{\x}$.
The loss function for this encoder-decoder network is the reconstruction error:
\begin{align}
\begin{split}
\mathcal{L}_0 = \frac{1}{2N} \sum_{j=1}^N \|\x_j - \hat \x_j \|^2_2  = \frac{1}{2N} \| X - \hat X \|_F^2,
\end{split}
\label{eq:CAE-cost-0}
\end{align}
where $N$ is the number of images in the training set.

\myparagraph{Self-Expression Module}
State-of-the-art subspace clustering methods are based on the self-expression property of data, which states that each data point in a union of subspaces can be expressed as a linear combination of other data points~\cite{Elham:CVPR09,Vidal:Springer16}. % as $X=XC,~~\text{diag}(C)=0$, where $C\in R^{N\times N}$ is the coefficient matrix.
In order to learn feature representations that are suitable for subspace clustering, we adopt a self-expression module that imposes the following loss function:
\begin{equation}
    \lambda \|C\|_\ell + \frac{1}{2} \|Z - ZC\|_F^2 ~~\text{s.t.}~~ \diag(C)=\0,
\label{eq:Self-expression}
\end{equation}
where $Z=\begin{bmatrix}\z_1,\cdots,\z_N \end{bmatrix}$ is a matrix containing features from the feature extraction module as its columns, $\|C\|_{\ell}$ is a properly chosen regularization term, % to yield a subspace-preserving solution,
the constraint $\text{diag}(C) = \0$ is optionally used to rule out a trivial solution of $C=I$, and $\lambda > 0$ %, $\lambda_2 > 0$ are
is a tradeoff parameter.

\myparagraph{Self-Supervision Modules}
Once the self-expression coefficient matrix $C$ is obtained,
we can compute a data affinity matrix as $A = \frac{1}{2}(|C|+|C^\top|)$.
Subsequently, spectral clustering can be applied on $A$ to obtain a segmentation of the data by minimizing the following cost:
%The data affinity $a_{ij}$ specifies a cost for segmenting data points $\z_i$ and $\z_j$ into two different groups. Thus, the clustering of the data is obtained by finding a segmentation of the data that minimizes the sum of such costs, \ie,
%
\begin{align}
\begin{split}
\min_{Q}~\sum_{i,j} a_{ij} \|\q_i - \q_j \|^2_2,~~\st~~Q \in \Q,
\end{split}
\label{eq:spectral-clustering-cost}
\end{align}
where $\Q = \{ Q \in \{0,1\}^{n \times N} : \1^\top Q = \1^\top ~ \text{and} ~ \rank(Q) = n \}$ is a set of all valid segmentation matrices with $n$ groups, and $\q_i$ and $\q_j$ are respectively the $i$-th and $j$-th columns of $Q$ indicating the membership of each data point to the assigned cluster. In practice, since the search over all $Q \in \Q$ is combinatorial, spectral clustering techniques usually relax the constraint $Q \in \Q$ to $QQ^\top =I$.

Observe that the spectral clustering produces a labeling of the data set which, albeit is not necessarily the correct class label for all the data points, contains meaningful information about the data.
This motivates us to supervise the training of the feature extraction and self-expression modules using the output of spectral clustering.
In principle, the features learned from the feature extraction module should contain enough information for predicting the class labels of the data points.
Therefore, we introduce a classification layer on top of the feature extraction module which is expected to produce labels that aligns with the labels generated in spectral clustering. % (with a proper permutation).
Furthermore, the segmentation produced by
spectral clustering can also be used to
construct a binary segmentation matrix, which contains information regarding which data points should be used in the expression of a particular data point.
Therefore, we incorporate the objective function of spectral clustering as a loss function in our network formulation, which has the effect of supervising the training of the self-expression module.
We present the details of these two self-supervision modules in the following two subsections.

\subsection{Self-Supervision for Self-Expression}
\label{sec:weak-supervision-self-express}

To exploit the information in the labels produced by spectral clustering, we incorporate spectral clustering as a module of the network which provides a feedback to the self-expression model (see Fig. \ref{fig:ConvSSCNet}).

To see how the objective function of spectral clustering in \eqref{eq:spectral-clustering-cost} provides such feedback, we rewrite \eqref{eq:spectral-clustering-cost} to a weighted $\ell_1$ norm of $C$ as in \cite{Li:CVPR15}, that is,
\begin{align}
\!\!\!\!\!\frac{1}{2} \sum_{i,j} a_{ij} \|\q_i - \q_j \|^2_2 = \sum_{i,j} |c_{ij}| \frac{\|\q_i - \q_j \|^2_2}{2} :=\|C \|_Q,
\label{eq:spectral-clu-cost-C-Q-norm}
\end{align}
where we have used the fact that $a_{ij} = \frac{1}{2}(|c_{ij}| + |c_{ji}|)$.
It can be seen from \eqref{eq:spectral-clu-cost-C-Q-norm} that $\|C\|_Q$ measures the discrepancy between the coefficients matrix $C$ and the segmentation matrix $Q$.
When $Q$ is provided, minimizing the cost $\|C\|_Q$ has the effect of enforcing the self-expression matrix $C$ to be such that an entry $c_{ij}$ is nonzero only if the $i$-th and $j$-th data points have the same class labels.
Therefore, incorporating the term $\|C\|_Q$ in the network formulation helps the training of the self-expression module. That is, the result of previous spectral clustering can be incorporated into the self-expression model to provide self-supervision for refining the self-expression matrix $C$.

\subsection{Self-Supervision for Feature Learning}
\label{sec:weak-supervision-convolutional-layer}

We also use the class labels generated from spectral clustering to supervise the training of the feature extraction module.
Notice that the output of spectral clustering is an $n$-dimensional vector which indicates the membership to $n$ subspaces (\ie, clusters). %\footnote{In this paper, we use two different ways to encode the membership to $n$ clusters: one is a binary indicator vector and another one is a real-valued vector.}.
Thus, we design FC layers as  $p \times N_1 \times N_2 \times n$, where $p$ is the dimension of the extracted convolutional feature, which is defined as the concatenation of the different feature maps of the last convolutional layer in the encoder block, and $N_1$ and $N_2 $ are the numbers of neurons in the two FC layers, respectively. % and $n$ is the number of subspaces.

Denote $\y$ as the $n$-dimensional output of the FC layers, where $\y \in \RR^n$.
Note that the output $\{\q_j\}_{j=1}^N$ of spectral clustering will be treated as %a temporal
the target output of the FC layers.
%To exploit the self-supervision information to train the convolutional encoder, we define a \textit{cross-entropy} (CE) loss as follows:
%%
%\begin{align}
%\begin{split}
%%\C_2 = - \frac{1}{N}\sum_{j=1}^N \ln \sigma (\tilde \y_j^\top \q_j),
%%\C_4 = \frac{1}{N}\sum_{j=1}^N \ln ( 1 + \exp(-\tilde \y_j^\top \q_j)).
%\L_4 =  \sum_{j=1}^N \ln ( 1 + \exp(-\tilde \y_j^\top \q_j)),
%\end{split}
%\label{eq:CAE-cost-Cross-Entropy}
%\end{align}
%%
%which is also denoted as $CE(\tilde Y, Q)$.
%%Note that $\{\q_j\}_{j=1}^N$ provides only a temporal target for the output of the FC layers.
%
To exploit the self-supervision information to train the convolutional encoder, we define a mixture of \textit{cross-entropy loss} and \textit{center loss} (CEC) as follows:
\begin{align}
\begin{split}
%\C_2 &= - \frac{1}{N}\sum_{j=1}^N \ln \sigma (\tilde \y_j^\top \q_j),\\
%\C_4 &= \frac{1}{N}\sum_{j=1}^N \ln ( 1 + \exp(-\tilde \y_j^\top \q_j)).\\
%\L_4 &=  \L_S+\lambda\L_C\\
\mathcal{L}_4 = \frac{1}{N}\sum_{j=1}^N (\ln ( 1 + \text{e}^{-\tilde \y_j^\top \q_j}) + \tau \| \y_j - {\bf \mu}_{\pi(\y_j)}\|^2_2),
\end{split}
\label{eq:CAE-cost-Cross-Entropy}
\end{align}
where $\tilde \y_j$ is a normalization of $\y_j$ via \emph{softmax},
% \ie,
%
%\begin{align}
%  \tilde y_{ij} = \frac{\exp(y_{ij})}{\sum_{i=1}^n \exp(y_{ij})},
%  \label{eq:soft-max}
%\end{align}
%
% to make the output of FC layers compatible with the output of spectral clustering,
$\bf{\mu}_{\pi(\y_j)}$ denotes the cluster center which corresponds to $\y_j$, $\pi(\y_j)$ is to take the index of $\y_j$ from the output of spectral clustering, and $0 \le \tau \le 1$ is a tradeoff parameter. The first term of  $\mathcal{L}_4$ is effectively a cross-entropy loss and the second term of $\mathcal{L}_4$ is a center loss which compresses the intra-cluster variations.
%The mixture of the cross-entropy loss and the center loss can improve the discriminative power of the learned feature. \cite{Wen:ECCV16}.

%The formulation $\L_C$ effectively characterizes the intra-cluster variations, which can improve the discriminative power of the learned feature \cite{Wen:ECCV16}.

%{\mcr{Recent work introduced an auxiliary function called the center loss, that increases}}
%The two FC layers have $p \times N_1 + N_1 \times N_2$ parameters, denoted as $\Theta^{(1)}$ and $\Theta^{(2)}$, which are needed to be effectively learned.

An important issue in defining such a loss function is that the output of spectral clustering $\{\q_j\}_{j=1}^N$ provides merely \emph{pseudo labels} for the input data. That is, %specific
the label index assigned to a cluster in the returned result of spectral clustering is up to an unknown permutation.
Therefore, the class labels from two successive epochs might not be consistent.
To address this issue, we propose to perform a permutation of the new pseudo labels via Hungarian algorithm \cite{Munkres:JSIAM57} to find an optimal assignment between the pseudo labels of successive iterations
%
% to make the new pseudo labels as consistent as possible
% so that it is most consistent with pseudo labels from the previous iteration
%
%
before feeding them into the self-supervision module with the cross-entropy loss in \eqref{eq:CAE-cost-Cross-Entropy}.

\myparagraph{Remark 1} Note that the output of spectral clustering is used in two interrelated self-supervision modules and thus we call it a \emph{dual self-supervision} mechanism.\footnote{
%{\mcr Owing to properly and effectively exploiting the self-supervision information, our proposed S$^2$ConvSCN yields superior performance.}
%
While it is also sensible to term our approach with ``self-training'', we prefer to use the term ``self-supervision'' in order to emphasizes on the mechanism of guiding the training of the whole framework, that is to make each component as consistent as possible (\ie, be separable, self-expressive, and block diagonal).}

\subsection{Training S$^2$ConvSCN}
\label{sec:training-ConvSSCNet}

To obtain an end-to-end trainable framework, we design the total cost function of S$^2$ConvSCN by putting together the costs %of the five components (\ie, $\mathcal{L}_0$, $\mathcal{L}_1$, $\mathcal{L}_2$, $\mathcal{L}_3$ and $\mathcal{L}_4$) as following:
in \eqref{eq:CAE-cost-0}, \eqref{eq:Self-expression}, \eqref{eq:spectral-clu-cost-C-Q-norm}, and \eqref{eq:CAE-cost-Cross-Entropy} as follows:
\begin{align}
\begin{split}
\mathcal{L} = \mathcal{L}_0 + \gamma_1 \mathcal{L}_1 + \gamma_2 \mathcal{L}_2 + \gamma_3 \mathcal{L}_3 + \gamma_4 \mathcal{L}_4,
%   &= \frac{1}{2N} \| X - \hat X \|_F^2 + \frac{1}{N}\sum_{j=1}^N \ln ( 1 + \exp(-\tilde \y_j^\top \q_j)) + \|C\|_{\ell} + \alpha \|C\|_{Q} + \lambda \|Z - Z C\|^2_{F}.
\end{split}
\label{eq:total-cost-ConvSSCNet}
\end{align}
where %$\mathcal{L}_0$ is defined in \eqref{eq:CAE-cost-0},
$\mathcal{L}_1 = \|C\|_{\ell}$,
$\mathcal{L}_2 = \frac{1}{2}\|Z - Z C\|^2_{F}$,
$\mathcal{L}_3 = \|C\|_Q$,
%$\mathcal{L}_4$ is defined in \eqref{eq:CAE-cost-Cross-Entropy},
and $\gamma_1$, $\gamma_2$, $\gamma_3$ and $\gamma_4$ are four tradeoff parameters.
The tradeoff parameters are set roughly to be inversely proportional to the value of each cost in order to obtain a balance amongst them.
%The effects on the subspace clustering performance of using different tradeoff parameters will be evaluated in experiments.

%\begin{align}
%%\begin{split}
%\C_1 &= \|C\|_{\ell}, \\
%\C_2 &= \frac{1}{2}\|Z - Z C\|^2_{F}.
%%\end{split}
%\label{eq:CAE-cost-Self-Express}
%\end{align}
%
%, and the cost function of the spectral clustering layer is:
%%
%\begin{align}
%\begin{split}
%\C_3 = \|C\|_Q.
%\end{split}
%\label{eq:CAE-cost-Spectral-cluster}
%\end{align}
%%
%%where $\sigma(t) = \frac{1}{1+\exp(-t)}$.

%{\mcb
%
\begin{algorithm}[tb]
\small
   \caption{Procedure for training S$^2$ConvSCN}
   \label{alg:ConvSSCNet-train}
\begin{algorithmic}[1]
      \REQUIRE \texttt{Input data, tradeoff parameters, maximum iteration T$_{\max}$, T$_0$, and t=1.} % criterion: ($\hat{\rho}$, $\hat{\C_0}$, $\hat{\C_2}$), and
      %\begin{enumerate}
      \STATE \texttt{Pre-train the stacked convolutional module via stacked CAE.} % while keeping other parts freezed.}
      \STATE \texttt{(Optional) Pre-train the stacked convolutional module with the self-expressive layer.} % while keeping other parts freezed.}
      \STATE \texttt{Initialize the FC layers.}
      \STATE \texttt{Run self-expressive layer.}
      \STATE \texttt{Run spectral clustering layer to get the segmentation Q.}
      \WHILE %{\texttt{ {(\mcr{$\rho \leq \hat{\rho}$ and $\C_0 \leq \hat{\C_0}$ and $\C_2 \leq \hat{\C_2}$})} or t $\leq$ T$_{\max}$ }} %\le
      {\texttt{ t $\leq$ T$_{\max}$ }} %\le
      \STATE \texttt{Fixed Q, update the other parts T$_0$ epoches.}
      \STATE \texttt{Run spectral clustering once to update Q and set t $\leftarrow$ t+1.}
      \ENDWHILE
      %\end{enumerate}
    \ENSURE \texttt{trained S$^2$ConvSCN and Q.}
\end{algorithmic}
\end{algorithm}

To train S$^2$ConvSCN, we propose a two-stage strategy as follows: a) %\textbf{Stage I}:
pre-train the stacked convolutional layers to provide an initialization of S$^2$ConvSCN;
b) train the whole network with the assistance of the self-supervision information provided by spectral clustering.

\myparagraph{Stage I: Pre-Training Stacked Convolutional Module} % Keep Other Layers Freezed
The pre-training stage uses the cost $\mathcal{L}_0$. In this stage, we set the weights in the two FC layers as zeros, %\ie, $\Theta^{(1)}=0$ and $\Theta^{(2)}=0$,
which yield zeros output. Meanwhile, we also set the output of spectral clustering as zero vectors, \ie, $\q_j=0$ for $j=1,\cdots,N$. By doing so, the two FC layers are ``sleeping'' during this pre-training stage.
%
% The stacked convolutional module is used as hierarchical feature extractor. We need to learn the parameters $\{\w_i^{(\ell)}\}_{i=1,\cdots,m}^{\ell=1,2,3}$.
%
Moreover, we set the coefficient matrix $C$ as an identity matrix, which is equivalent to training S$^2$ConvSCN without the self-expression layer. % $c_{ij}=\delta_{ij}$ to train the CAE.
As an optional pre-training, we can also use the pre-trained stacked CAE to train the stacked CAE with the self-expression layer. %This is in fact a DSCNet, which is investigated in \cite{Ji:NIPS17}.

\myparagraph{Stage II: Training the Whole S$^2$ConvSCN}
In this stage, we use the total cost $\mathcal{L}$ to train the whole S$^2$ConvSCN as a stacked CAE assisted with the self-expression module and dual self-supervision. %auxiliary components to exploit the self-supervision information.
To be more specific, given the spectral clustering result $Q$, we update the other parameters in S$^2$ConvSCN for $T_0$ epoches, and then perform spectral clustering to update $Q$. For clarity, we provide the detailed procedure to train S$^2$ConvSCN in Algorithm \ref{alg:ConvSSCNet-train}.

\myparagraph{Remark 2} In the total cost function as \eqref{eq:total-cost-ConvSSCNet}, if we set $\gamma_3=\gamma_4=0$, then the two self-supervision blocks will disappear and our S$^2$ConvSCN reduces to DSCNet \cite{Ji:NIPS17}. Thus, it would be interesting to add an extra pre-training stage, \ie, using the cost function $\mathcal{L}_0 + \gamma_1 \mathcal{L}_1 + \gamma_2 \mathcal{L}_2$ to train the stacked convolutional module and the self-expressive layer together before evoking the FC layers and the spectral clustering layer. This is effectively a DSCNet \cite{Ji:NIPS17}.
In experiments, as used in \cite{Ji:NIPS17}, we stop the training by setting a maximum number of epoches $T_{\max}$.

\begin{table}[t]
%\vskip -0.1in
%\vskip -0.12in
\center
\small
%\begin{center}
%\begin{small}
%\begin{sc}
%\resizebox{0.65\textwidth}{!}{ %0.95\columnwidth
\begin{tabular}{c | cc | cc }
    %\toprule
    \hline
     & \multicolumn{2}{c|}{\textbf{Extended Yale B}} & \multicolumn{2}{c}{\textbf{ORL}}\\
    %\midrule
    \hline
    Layers & kernel size & channels & kernel size & channels \\
    %\midrule
    \hline
    encoder-1 & $5\times 5$ & 10 & $3\times 3$ & 3 \\
    encoder-2 & $3\times 3$ & 20 & $3\times 3$ & 3 \\
    encoder-3 & $3\times 3$ & 30 & $3\times 3$ & 5 \\
    decoder-1 & $3\times 3$ & 30 & $3\times 3$ & 5 \\
    decoder-2 & $3\times 3$ & 20 & $3\times 3$ & 3 \\
    decoder-3 & $5\times 5$ & 10 & $3\times 3$ & 3 \\
    %\bottomrule
    \hline
\end{tabular}
%}
%\end{sc}
%\end{small}
%\end{center}
%\vskip -0.15in
\caption{Network settings for Extended Yale B and ORL.}
\label{tab:Network-settings-ExYaleB-ORL-COIL20}
\end{table}

\begin{table*}[t]
\scriptsize
\centering
%\begin{tabular}{c|p{0.50cm}<{\centering}p{0.50cm}<{\centering}p{0.50cm}<{\centering}p{0.50cm}<{\centering}p{0.50cm}<{\centering}
%p{0.50cm}<{\centering}p{0.50cm}<{\centering}p{0.65cm}<{\centering}
%p{0.75cm}<{\centering}p{0.75cm}<{\centering}
%|p{0.95cm}<{\centering}p{0.95cm}<{\centering}} % c
\begin{tabular}{c|c c c c c c c c c c c |c c}
\hline
%Methods & LRR & LRSC & LSR & SSC & AE + SSC & KSSC & SSC-OMP & EDSC & AE + EDSC & DSCNet-$\ell_1$ & DSCNet-$\ell_2$
%& ConvSC-Hard-$\ell_2$ & ConvSC-Hard-$\ell_1$ & ConvSC-Soft-$\ell_2$ & ConvSC-Soft-$\ell_1$ \\ %&ConvSSCNet(a)\footnote{In ConvSSCNet(a), we replace $\|C\|_\ell$ by $\1^\top C \1$.} \\
\!Methods & \!LRR & \!LRSC & \!SSC & \!\!AE+ SSC & \!KSSC & \!SSC-OMP & \!soft S$^3$C$^\dag$ & \!EDSC & \!AE+ EDSC & \!DSC-$\ell_1$ & \!DSC-$\ell_2$
& {\bf Ours} ($\ell_2$) & {\bf Ours} ($\ell_1$) \\
%& \!\!S$^2$ConvSCN-$\ell_2$ & \!\!S$^2$ConvSCN-$\ell_1$ \\
%
%
%& \!\!S$^2$ConvSCN-$\ell_2$(soft) & \!\!S$^2$ConvSCN-$\ell_1$(soft) \\ %&ConvSSCNet(a)\footnote{In ConvSSCNet(a), we replace $\|C\|_\ell$ by $\1^\top C \1$.} \\
\hline
\multicolumn{12}{l}{\textbf{10 subjects}} \\
\hline
Mean   &19.76 &30.95 & 8.80 &17.06& 14.49& 12.08   & 6.34 & 5.64 &5.46& 2.23& 1.59& \textbf{1.18} & \textbf{1.18} \\ %& 1.56 & 1.56 \\
Median &18.91 &29.38 & 9.06& 17.75& 15.78& 8.28 & 3.75 &5.47& 6.09& 2.03& 1.25& \textbf{1.09} & \textbf{1.09} \\% & 1.25 & 1.25 \\
\hline
\multicolumn{12}{l}{\textbf{15 subjects}} \\
\hline
Mean   &25.82  &31.47 & 12.89  & 18.65 &16.22 &14.05& 11.01    & 7.63 &6.70 &2.17 &1.69& \underline{1.14}& \textbf{1.12} \\%& 1.33 & 1.32 \\
Median &26.30  &31.64 & 13.23  &17.76  &17.34 &14.69& 10.89    & 6.41 &5.52 &2.03 &1.72& \textbf{1.14}& \textbf{1.14} \\%& 1.30 & 1.25 \\
\hline
\multicolumn{12}{l}{\textbf{20 subjects}}\\
\hline
Mean  &31.45 &28.76 &20.11& 18.23& 16.55& 15.16   & 14.07         &9.30 &7.67& 2.17 &1.73& \underline{1.31}&\textbf{1.30} \\%& 1.24 & 1.27 \\
Median&32.11 &28.91 &21.41& 16.80& 17.34 &15.23 & 13.98         &10.31& 6.56 &2.11 &1.80&\underline{1.32}&\textbf{1.25} \\% & 1.33 & 1.33 \\
\hline
\multicolumn{12}{l}{\textbf{25 subjects}} \\
\hline
Mean  &28.14& 27.81& 26.30 &18.72& 18.56 &18.89   &16.79      &10.67 &10.27 &2.53 &1.75& \underline{1.32}&\textbf{1.29} \\ %& 1.36 & 1.36 \\
Median&28.22& 26.81& 26.56 & 17.88& 18.03 &18.53 &17.13      &10.84& 10.22 &2.19 &1.81& \underline{1.34}&\textbf{1.28} \\% & 1.34 &1.34 \\
\hline
\multicolumn{12}{l}{\textbf{30 subjects}} \\
\hline
Mean  &38.59 &30.64& 27.52 &19.99& 20.49& 20.75   & 20.46     &11.24& 11.56 &2.63 &2.07& \underline{1.71}&\textbf{1.67} \\ %& 1.58 & 1.59 \\
Median&36.98 &30.31& 27.97& 20.00 &20.94 &20.52 & 21.15     &11.09 &10.36 &2.81 &2.19& \underline{1.77} &\textbf{1.72} \\ %& 1.61 & 1.61 \\
\hline
\multicolumn{12}{l}{\textbf{35 subjects}}\\
\hline
Mean   &40.61&31.35& 29.19& 22.13 &26.07& 20.29   &20.38      &13.10& 13.28  & 3.09& 2.65&\underline{1.67}&\textbf{1.62} \\ %& 2.12 & 2.11 \\
Median &40.71&31.74& 29.51& 21.74& 25.92& 20.18&20.47       &13.10& 13.21& 3.10& 2.64& \underline{1.69}&\textbf{1.60} \\ %& 2.14 & 2.01 \\
\hline
\multicolumn{12}{l}{\textbf{38 subjects}} \\
\hline
Mean   &35.12 &29.89 & 29.36& 25.33& 27.75   & 23.52  &19.45      &11.64 &12.66 &3.33& 2.67 & \underline{1.56} & \textbf{1.52} \\ %& 2.75 & 2.75 \\
Median &35.12 &29.89 & 29.36& 25.33& 27.75& 23.52 &19.45      & 11.64 &12.66& 3.33& 2.67 & \underline{1.56} &\textbf{1.52} \\ %& 2.75 & 2.75 \\
\hline
\end{tabular}
\caption{Clustering Error (\%) on Extended Yale B. The best results are in bold and the second best results are underlined.} %In ConvSSCNet(a), we replace $\|C\|_\ell$ by $\1^\top C \1$. }
%\vspace{-10pt}
\label{table:extendedYaleB-DS3C}
\end{table*}

\section{Experimental Evaluations}
\label{sec:experiments}

To evaluate the performance of our proposed S$^2$ConvSCN, we conduct experiments on four benchmark data sets: two face image data sets, the Extended Yale B \cite{Georghiades:TPAMI01} and ORL \cite{Sam:94}, and two object image data sets, COIL20 and COIL100~\cite{Nene:COIL}.
%
%\myparagraph{Datasets Description}
%
%
%\myparagraph{Baselines}
%
We compare our proposed S$^2$ConvSCN with the following baseline algorithms, including Low Rank Representation (LRR)~\cite{Liu:ICML10}, Low Rank Subspace Clustering (LRSC) \cite{ReVidal:PRL14},
%
%{\mcr Least Square Regression (LSR) \cite{Lu:ECCV12},}
%
Sparse Subspace Clustering (SSC)~\cite{Elham:TPAMI13}, Kernel Sparse Subspace Clustering (KSSC) \cite{Patel:ICIP14}, SSC by Orthogonal Matching Pursuit (SSC-OMP)~\cite{You:CVPR16}, Efficient Dense Subspace Clustering (EDSC) \cite{Ji:WACV14},
Structured SSC (S$^3$C) \cite{Li:TIP17}, SSC with the pre-trained convolutional auto-encoder features (AE+SSC), EDSC with the pre-trained convolutional auto-encoder features (AE+EDSC), Deep Subspace Clustering Networks (DSCNet) \cite{Ji:NIPS17} and Deep Adversarial Subspace Clustering (DASC)~\cite{Zhou:CVPR18}. For EDSC, AE+EDSC, DSCNet and DASC, we directly cite the best results reported in \cite{Ji:NIPS17} and \cite{Zhou:CVPR18}. For S$^3$C, we use {\em soft} S$^3$C with a fixed parameter $\alpha=1$.

The architecture specification of S$^2$ConvSCN used in our experiments for each dataset are listed in Table \ref{tab:Network-settings-ExYaleB-ORL-COIL20} and Table~\ref{table:setting for COIL}. % (COIL20 and COIL100).
In the stacked convolutional layers, we set the kernel stride as 2 in both horizontal and vertical directions, and use Rectified Linear Unit (ReLU) \cite{Krizhevsky:NIPS2012} as the activation function $\sigma(\cdot)$. In addition, the learning rate is set to $1.0 \times 10^{-3}$ in all our experiments. The whole data set is used as one batch input. % to minimize the cost.
%{\mcr We set $N_1 = ?$ and $N_2 = ?$.}
For the FC layers, we set $N_1 =\frac{N}{2}$ and $N_2 =n$.

To find informative affinity matrix, we adopt the vector $\ell_1$ norm and the vector $\ell_2$ norm to define $\| C \|_\ell$ and denote as S$^2$ConvSCN-$\ell_1$ and S$^2$ConvSCN-$\ell_2$, respectively. In the second training stage, we update the stacked convolutional layers, the self-expression model, and the FC layers for $T_0$ epochs and then update the spectral clustering module once, where $T_0$ is set to $5\sim 16$ in our experiments. % $\{10, 25, 5, 50\}$ in our experiments. % to 50.

\subsection{Experiments on Extended Yale B}

The Extended Yale B database \cite{Georghiades:TPAMI01} consists of face images of 38 subjects, 2432 images in total, with approximately 64 frontal face images per subject taken under different illumination conditions, where the face images of each subject correspond to a low-dimensional subspace. In our experiments, we follow the protocol used in %\cite{Elham:TPAMI13} and
\cite{Ji:NIPS17}: a) each image is down-sampled from $192\times 168$ to $48 \times 42$ pixels; b) experiments are conducted using all choices of $n\in \{10,15,20,25,30,35,38\}$. % subjects.

To make a fair comparison, we use the same setting as that used in DSCNet \cite{Ji:NIPS17}, in which a three-layer stacked convolutional encoders is used with $\{10, 20, 30\}$ channels, respectively. The detailed settings for the stacked convolutional network used on Extended Yale B are shown Table \ref{tab:Network-settings-ExYaleB-ORL-COIL20}.
The common parameters $\gamma_1$ and $\gamma_2$ are set the same as that in DSCNet, where $\gamma_1=1$ (for the term $\|C\|_\ell$) and $\gamma_2=1.0\times 10^{\frac{n}{10}-3}$. %$\lambda_1=1.0$, $\lambda_2=1.0\times 10^{\frac{n}{10}-3}$.
For the specific parameters used in S$^2$ConvSCN, we set $\gamma_3=16$ for the term $\|C\|_Q$ and %$\gamma_4=2^{10}$
$\gamma_4=72$ for the cross-entropy term, respectively.  %we set $\alpha=20$ for subspace structured norm \eqref{Q_norm} and $\beta=100$ for cross-entropy term.
We set $T_0=5$ and $T_{\max}=10+40n$. % where $n$ is the number of clusters.

The experimental results are presented in Table~\ref{table:extendedYaleB-DS3C}. We observe that our proposed S$^2$ConvSCN-$\ell_1$ and S$^2$ConvSCN-$\ell_2$ remarkably reduced the clustering errors and yield the lowest clustering errors with $n\in \{10,15,20,25,30,35,38\}$ than all the listed baseline methods. %; whereas for $n=38$, the clustering errors of S$^2$ConvSCN-$\ell_1$ and S$^2$ConvSCN-$\ell_2$ are still the second best. %lower than the DSC-Net-L1, as get a second place total
We note that DASC \cite{Zhou:CVPR18} reported a clustering error of $1.44\%$ on Extended Yale B with $n=38$, which is slightly better than our results.

%\myparagraph{Evaluation on Dual Self-Supervision}
To gain further understanding of the proposed dual self-supervision, we use S$^2$ConvSCN-$\ell_1$ as an example and evaluate the effect of using the dual self-supervision modules via an ablation study. Due to space limitation, we only list the experimental results of using a single self-supervision via $\mathcal{L}_3$, using a single self-supervision via $\mathcal{L}_4$, and using dual self-supervision of $\mathcal{L}_3$ plus $\mathcal{L}_4$ on datasets Extended Yale B in Table \ref{table:ablation-study-extendedYaleB-DS3C}. % and ORL
As a baseline, we show the experimental results of DSCNet \cite{Ji:NIPS17}, which uses the loss $\mathcal{L}_0 + \mathcal{L}_1 + \mathcal{L}_2$. As could be read from Table \ref{table:ablation-study-extendedYaleB-DS3C} that, using only a single self-supervision module, \ie, $\mathcal{L}_0 + \mathcal{L}_1 + \mathcal{L}_2$ plus $\mathcal{L}_3$, or $\mathcal{L}_0 + \mathcal{L}_1 + \mathcal{L}_2$ plus $\mathcal{L}_4$, the clustering errors are reduced. Compared to using the self-supervision via a spectral clustering loss $\mathcal{L}_3$ in the self-expression module, using the self-supervision via the classification loss $\mathcal{L}_4$ in FC block is more effective. Nonetheless, using the dual supervision modules further reduces the clustering errors.

\begin{table*}[t]
%\caption{$S^2ConvSCN-l_1$.}
\centering
\resizebox{2.08\columnwidth}{!}
{
\begin{tabular}{c|cc|cc|cc|cc|cc|cc|cc} % c
\hline
\multirow{2}{*}{\diagbox{\textbf{Losses}}{\textbf{No. Subjects}}} & \multicolumn{2}{c|}{\textbf{10 subjects}} & \multicolumn{2}{c|}{\textbf{15 subjects}} & \multicolumn{2}{c|}{\textbf{20 subjects}} & \multicolumn{2}{c|}{\textbf{25 subjects}} & \multicolumn{2}{c|}{\textbf{30 subjects}} & \multicolumn{2}{c|}{\textbf{35 subjects}} & \multicolumn{2}{c}{\textbf{38 subjects}} \\
\cline{2-15}
 & Mean & Median & Mean & Median & Mean & Median & Mean & Median & Mean & Median & Mean & Median & Mean & Median \\
\hline
$\mathcal{L}_0+\mathcal{L}_1+\mathcal{L}_2$(DSC-$\ell_1$ \cite{Ji:NIPS17}) & 2.23 & 2.03 & 2.17 & 2.03 & 2.17 & 2.11 & 2.53 & 2.19 & 2.63 & 2.81 & 3.09 & 3.10 & 3.33 & 3.33 \\
\hline
$\mathcal{L}_0+\mathcal{L}_1+\mathcal{L}_2+\mathcal{L}_3$ & 1.58 & 1.25 & 1.63 & 1.55 & 1.67 & 1.57 & 1.61 & 1.63 & 2.74 & 1.82 & 2.64 & 2.65 & 2.75 & 2.75 \\
$\mathcal{L}_0+\mathcal{L}_1+\mathcal{L}_2+\mathcal{L}_4$ & 1.32 & \textbf{1.09} & 1.31 & 1.30 & 1.54 & 1.48 & 1.48 & 1.98 & 1.87 & \textbf{1.61} & 1.82 & 1.84 & 1.92 & 1.92\\
\hline
$\mathcal{L}_0+\mathcal{L}_1+\mathcal{L}_2+\mathcal{L}_3+\mathcal{L}_4$ & \textbf{1.18} & \textbf{1.09} & \textbf{1.12} & \textbf{1.14} & \textbf{1.30} & \textbf{1.25} & \textbf{1.29} & \textbf{1.28} & \textbf{1.67} & {1.72} & \textbf{1.62} & \textbf{1.60} & \textbf{1.52} & \textbf{1.52} \\
\hline

\end{tabular}
}
\caption{Ablation Study on S$^2$ConvSCN-$\ell_1$ on Extended Yale B.}
\label{table:ablation-study-extendedYaleB-DS3C}
%\vspace{-10pt}
\end{table*}

\subsection{Experiments on ORL}

The ORL dataset \cite{Sam:94} consists of face images of 40 distinct subjects, each subjects having 10 face images under varying lighting conditions, with different facial expressions (open/closed eyes, smiling/not smiling) and facial details (glasses / no glasses) \cite{Sam:94}. As the images were took under variations of facial expressions, this data set is more challenging for subspace clustering due to the %each subspace is more
nonlinearity and small sample size per subject. % and it contains 400 images totally compared with 2432 images in Extended Yale B.

In our experiments, each image is down-sampled from $112\times 92$ to $32\times 32$. We reduce the kernel size in convolution module to $3 \times 3$ due to small image size and set the number of channels to $\{3, 3, 5\}$. The specification of the network structure is shown in Table \ref{tab:Network-settings-ExYaleB-ORL-COIL20}. %Due to it contains only 400 images,
For the tradeoff parameters, we set $\gamma_1 =0.1, \gamma_2 =0.01, \gamma_3 =8$, and $\gamma_4 =1.2$ for our S$^2$ConvSCN. For the fine-tuning stage, we set $T_0 =5$ and %the number of maximum epochs
$T_{\max}=940$.
Experimental results are shown in Table~\ref{Table:acc-for-orl-and-coil}. Again, our proposed approaches %S$^2$ConvSCN-$\ell_1$ and S$^2$ConvSCN-$\ell_2$
yield the best results.

\subsection{Experiments on COIL20 and COIL100}

To further verify the effectiveness of our proposed S$^2$ConvSCN, we conduct experiments on dataset COIL20 %~\cite{Nene:COIL20}
and COIL100~\cite{Nene:COIL}. COIL20 contains 1440 gray-scale images of 20 objects; whereas COIL100 contains 7200 images of 100 objects. Each image was down-sampled to $32\times 32$. The settings of the stacked convolutional networks used for COIL20 and COIL100 are listed in Table~\ref{table:setting for COIL}.

For the tradeoff parameters on COIL20, we set $\gamma_1 =1$, $\gamma_2 =30$ as same as used in DSC-Net~\cite{Ji:NIPS17}, and %$\gamma_3 =8$, $\gamma_4 =2^{9}$,
$\gamma_3=8$, $\gamma_4=6$, $T_0 =4$, and $T_{\max}=80$ in our S$^2$ConvSCN.
For the tradeoff parameters on COIL100, we set $\gamma_1 =1$, $\gamma_2 =30$ as same as used in DSC-Net~\cite{Ji:NIPS17},
% and $\gamma_3 =128$, $\gamma_4 =1024$, $T_0 =16$, and $T_{\max}=180$ in our S$^2$ConvSCN.
and $\gamma_3 =8$, $\gamma_4 =7$, $T_0 =16$, and $T_{\max}=110$ in our S$^2$ConvSCN.

% To make a fair comparison to DSCNet \cite{Ji:NIPS17}
For experiments on COIL20 and COIL100, we initialize the convolutional module with stacked CAE at first, and then train a stacked CAE assisted with a self-expressive model. This is effectively DSCNet \cite{Ji:NIPS17}. And then, we train the whole S$^2$ConvSCN. Experimental results are listed in Table \ref{Table:acc-for-orl-and-coil}. As could be read, our S$^2$ConvSCN-$\ell_1$ and S$^2$ConvSCN-$\ell_2$ reduce the clustering errors significantly. This result confirms the effectiveness of the designed dual self-supervision components for the proper use of the useful information from the output of spectral clustering.

\begin{figure}
\vspace{-10pt}
\small
\centering
\subfigure[$\mathcal{L}, \mathcal{L}_0$ and $\mathcal{L}_2$]{\includegraphics[clip=true,trim=0 5 5 2, width=0.156\textwidth]{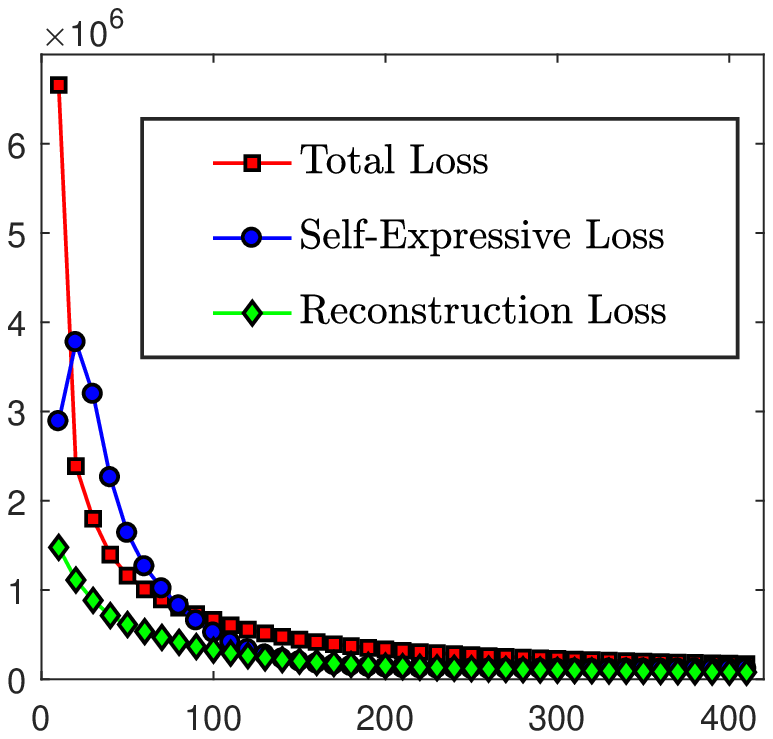}}
\subfigure[$\mathcal{L}_1$ and $\mathcal{L}_3$]{\includegraphics[clip=true,trim=0 5 5 2, width=0.156\textwidth]{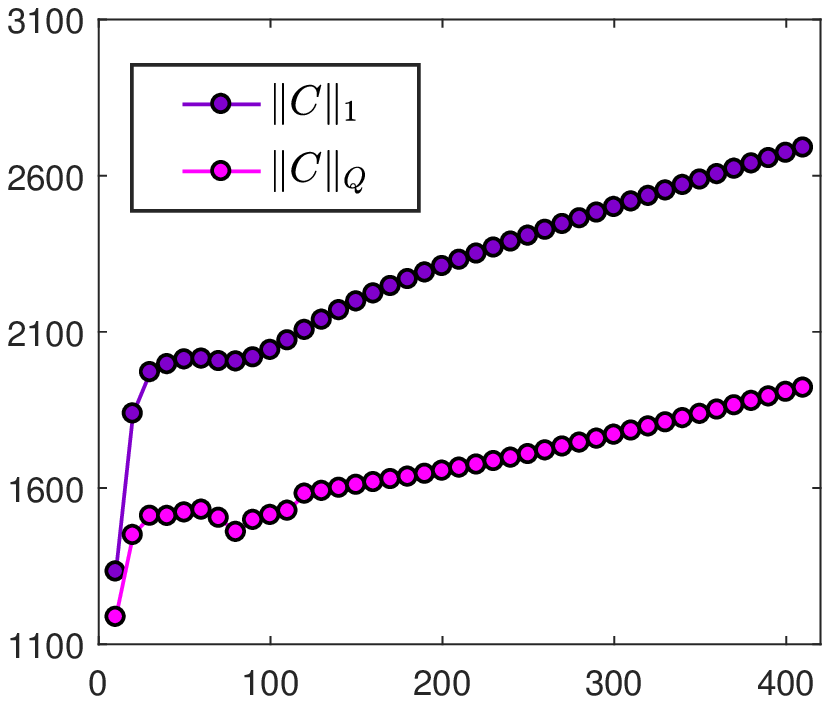}}
\subfigure[$\mathcal{L}_4$]{\includegraphics[clip=true,trim=0 5 5 2, width=0.156\textwidth]{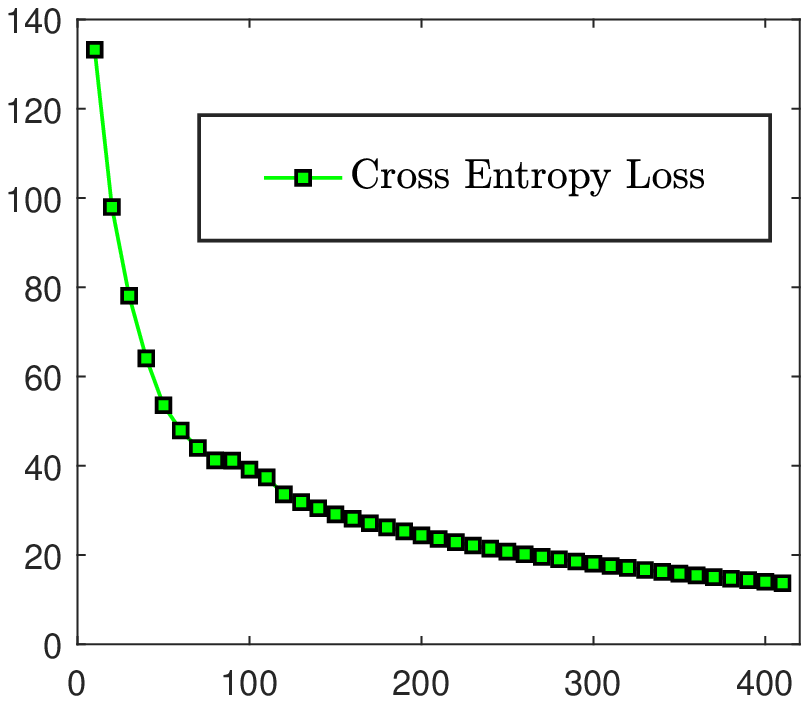}}\\
\subfigure[$\mathcal{L}_4$]{\includegraphics[clip=true,trim=0 5 5 2, width=0.156\textwidth]{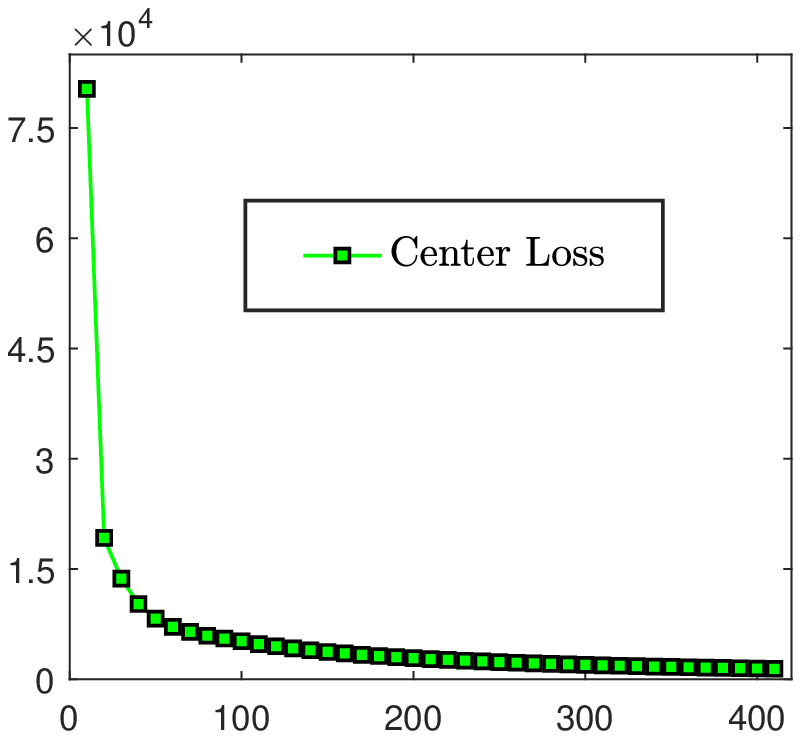}}
\subfigure[]{\includegraphics[clip=true,trim=0 5 5 2, width=0.156\textwidth]{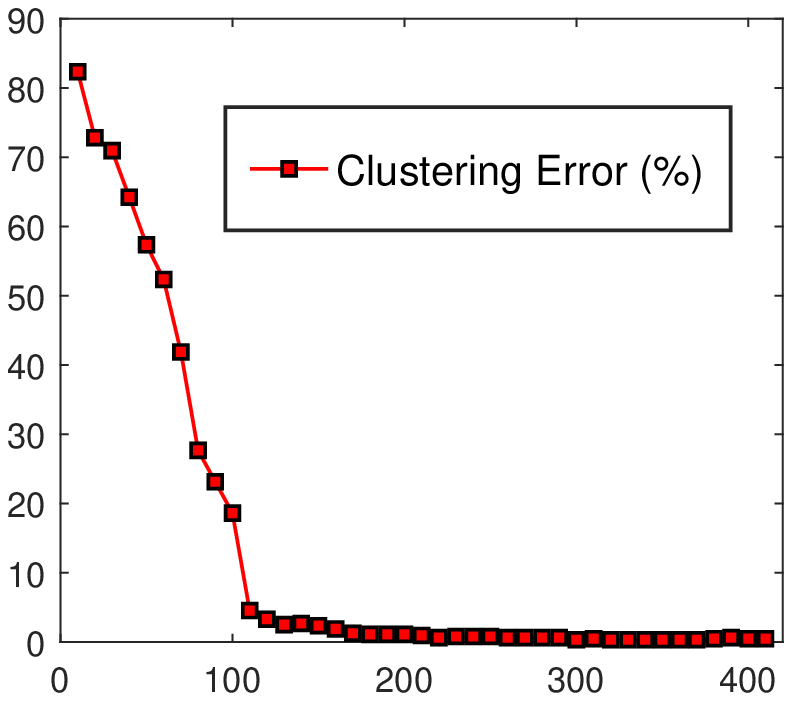}} % Clustering Error (\%)
\subfigure[$\frac{\mathcal{L}_3}{\mathcal{L}_1}$]{\includegraphics[clip=true,trim=0 5 5 2, width=0.156\textwidth]{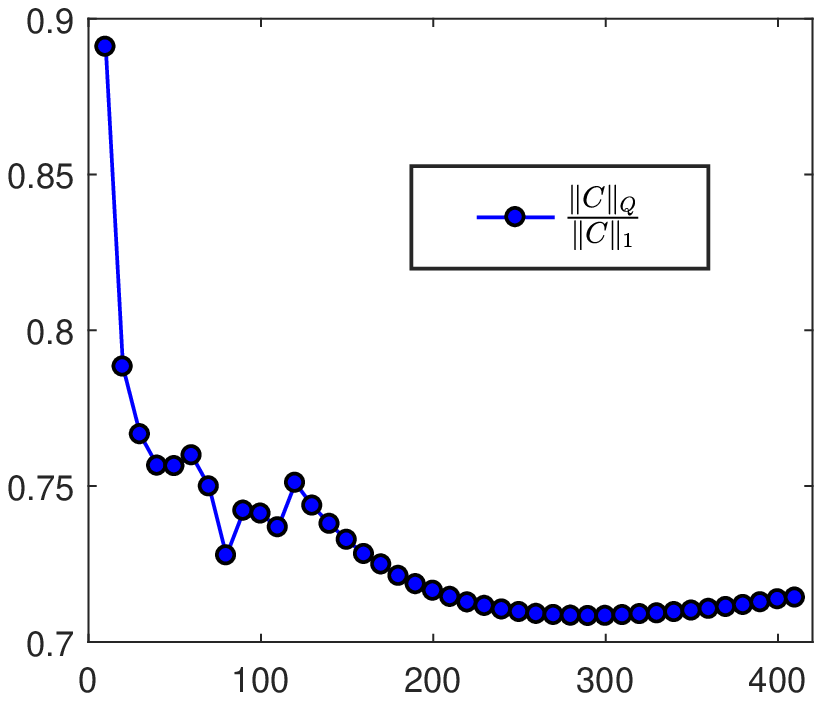}}
%\vskip -0.2in [clip=true,trim=5 0 5 0,width=0.27\columnwidth]
\caption{The cost functions and clustering error of S$^2$ConvSCN-$\ell_1$ during training period on Extended Yale B ($n=10$). }
\label{fig:Acc-Costs-vs-epoch-ExYaleB}
%\vspace{-2pt}
\end{figure}

\subsection{Convergence Behaviors}

To show the convergence behavior during training iterations, we conduct experiments on Extended Yale B with $n=10$. We record the clustering errors and each cost function during training period, and show them as a function of the number of epoches in Fig.~\ref{fig:Acc-Costs-vs-epoch-ExYaleB}. %we show the clustering accuracy and the used loss functions during training period, as a function of the number of epoches.
As could be observed from Fig.~\ref{fig:Acc-Costs-vs-epoch-ExYaleB}(a), (c), (d) and (e), the cost functions $\mathcal{L}$, $\mathcal{L}_0$, $\mathcal{L}_2$,
and $\mathcal{L}_4$, and the cluster error decrease rapidly and tend to ``flat''. To show more details in the iterations, in Fig.~\ref{fig:Acc-Costs-vs-epoch-ExYaleB} (b) and (f), %we observe a slightly ``elbow'' shape in the cost function $\C_2$ in the self-expressive layer and the cost function $\C_3$ in the spectral clustering layer. Interestingly, as shown in Figure \ref{fig:Acc-Costs-vs-epoch-ExYaleB}(f),
we show the curves of $\|C\|_1$, $\|C\|_Q$ and $\frac{\|C\|_Q}{\|C\|_1}$. Note that $\|C\|_Q$ and $\frac{\|C\|_Q}{\|C\|_1}$ are the cost and the relative cost of spectral clustering, respectively. Compared to $\|C\|_Q$, we argue that $\frac{\|C\|_Q}{\|C\|_1}$ is more indicative to the clustering performance. As could be observed, while $\|C\|_1$ and $\|C\|_Q$ are increasing\footnote{The observation that the curves of $\mathcal{L}_1$ and $\mathcal{L}_3$ go up is because the entries of the extracted feature $Z$ are slowly shrinking and thus the absolute values of entries of $C$ are slowly increasing, due to the absence of normalization step in feature learning at each epoch.}, the curve of $\frac{\|C\|_Q}{\|C\|_1}$ tends to ``flat''---which is largely consistent to the curve of the clustering error in Fig.~\ref{fig:Acc-Costs-vs-epoch-ExYaleB} (e).

\begin{table}[t]
\centering
\begin{tabular}{c|cc|cc}
    %\toprule
    \hline
    & \multicolumn{2}{c|}{COIL20} & \multicolumn{2}{c}{COIL100}\\
%    \midrule
 \hline
    Layers & kernel size & channels & kernel size & channels\\
    \hline %\midrule
    encoder-1 & $3\times 3$ & 15 & $5\times 5$ & 50\\
    decoder-1 & $3\times 3$ & 15 & $5\times 5$ & 50\\
\hline %    \bottomrule
\end{tabular}
%\end{sc}
%}
%\end{small}
%\end{center}
%\vskip -0.1in
\caption{Network settings for COIL20 and COIL100.} % COIL10 and
\label{table:setting for COIL}
%\vspace{-15pt}
\end{table}

%-------------------------------------------------------------
\begin{table}[htbp]
\centering
%\setlength{\abovecaptionskip}{0.cm}
%\setlength{\belowcaptionskip}{10pt}
%\small
\scriptsize
\resizebox{0.450\textwidth}{!}{
\begin{tabular}{c|ccc}
    \hline
    Methods & \text{ORL} & \text{COIL20} & \text{COIL100}\\
    \hline
% old: LRR & 38.25 & 31.01 & 59.82 \\
    LRR & 33.50 & 30.21 & 53.18 \\ % NEW
    LRSC & 32.50 & 31.25 & 50.67 \\
%   LSR1 & 18.50 & 41.11 & 53.30 \\ % [0,1] preprocessed by scaling, NOT PCA
%    LSR & 22.18 &22.60 & 54.19 \\ % preprocessed by L2 normalization
% old:    AE+SSC & 26.75 & 22.08 & 43.93 \\
% old:    SSC & 32.50 & 14.86 & 45.00 \\
    SSC & 29.50 & 14.83 & 44.90 \\ % NEW
    AE+SSC & 26.75 & 22.08 & 43.93 \\
    KSSC & 34.25 & 24.65 & 47.18 \\
% old:    SSC-OMP & 36.00 & 45.90 & 66.39 \\
    SSC-OMP & 37.05 & 29.86 & 67.29 \\ % NEW
%    SSC & 25.68 & 14.79 & 44.90 \\ % NEW 2 by Li  alpha =200
    EDSC & 27.25 & 14.86 & 38.13 \\
    AE+EDSC & 26.25 & 14.79 & 38.88 \\
%    soft S$^3$C$^\dag$ & 25.75 & 12.65 & 30.42 \\ % OLD
    soft S$^3$C$^\dag$ & 26.00 & 11.87 & 41.71 \\ % NEW
\hline
    DSC-$\ell_1$ & 14.25 & 5.65 & 33.62 \\
    DSC-$\ell_2$ & 14.00 & 5.42 & 30.96 \\
    DASC~\cite{Zhou:CVPR18} & 11.75 & 3.61 & - \\
    \hline
    S$^2$ConvSCN-$\ell_2$ & \underline{11.25} & \underline{2.33} & \underline{27.83}\\
    S$^2$ConvSCN-$\ell_1$ & \textbf{10.50} & \textbf{2.14} & \textbf{26.67} \\
    \hline
\end{tabular}
}
\caption{Clustering Error (\%) on ORL, COIL20 and COIL100.} % The best results are in bold. The second best results are underlined.}
\label{Table:acc-for-orl-and-coil}
\end{table}

\section{Conclusion}
\label{sec:conclusion}
We have proposed an end-to-end trainable framework for simultaneous feature learning and subspace clustering, called Self-Supervised Convolutional Subspace Clustering Network (S$^2$ConvSCN). Specifically, in S$^2$ConvSCN, the feature extraction via stacked convolutional module, the affinity learning via self-expression model, and the data segmentation via spectral clustering are integrated into a joint optimization framework. By exploiting a dual self-supervision mechanism, the output of spectral clustering are effectively used to improve the training of the stacked convolutional module and to refine the self-expression model, leading to superior performance. Experiments on benchmark datasets have validated the effectiveness of our proposed approach.

\section*{Acknowledgment} J. Zhang and C.-G. Li are supported by the National Natural Science Foundation of China (NSFC) under Grant No. 61876022, and the Open Project Fund from Key Laboratory of Machine Perception (MOE), Peking University. H. Zhang is partially supported by NSFC under Grant Nos. 61701032 and 61806184. X. Qi is supported by Shenzhen Fundamental Research Fund under Grants Nos. ZDSYS201707251409055 and 2017ZT07X152. Z. Lin is supported by 973 Program of China under Grant No. 2015CB352502, NSFC under Grant Nos. 61625301 and 61731018, Qualcomm, and Microsoft Research Asia.

%J. Zhang and C.-G. Li are supported by the National Natural Science Foundation of China (NSFC) under Grant No. 61876022, and the Open Project Fund from Key Laboratory of Machine Perception (MOE), Peking University. H. Zhang is partially supported by NSFC (Grant Nos. 61701032 and 61806184). Z. Lin is supported by National Basic Research Program of China (Grant No.: 2015CB352502), NSFC (Grant Nos. 61625301 and 61731018), and Microsoft Research Asia.

%This work was partially supported by the National Natural Science Foundation of China under Grant No. 61876022 and the Open Project Fund from Key Laboratory of Machine Perception (MOE), Peking University.

%\section*{References}
\small
\bibliographystyle{ieee} % {ieee_fullname}%
\bibliography{zhjj,zhjj_v1,cgli,temp,sparse,learning,vidal,recognition} % deeplearning

\begin{thebibliography}{10}\itemsep=-1pt

\bibitem{Dalal:CVPR05}
N.~Dalal and B.~Triggs.
\newblock Histograms of oriented gradients for human detection.
\newblock In {\em {IEEE} Conference on Computer Vision and Pattern
  Recognition}, 2005.

\bibitem{Elham:CVPR09}
E.~Elhamifar and R.~Vidal.
\newblock Sparse subspace clustering.
\newblock In {\em Proceedings of {IEEE} International Conference on Computer
  Vision and Pattern Recognition}, pages 2790--2797, 2009.

\bibitem{Elham:TPAMI13}
E.~Elhamifar and R.~Vidal.
\newblock Sparse subspace clustering: Algorithm, theory, and applications.
\newblock {\em {IEEE} Transactions on Pattern Analysis and Machine
  Intelligence}, 35(11):2765--2781, 2013.

\bibitem{Favaro:CVPR11}
P.~Favaro, R.~Vidal, and A.~Ravichandran.
\newblock A closed form solution to robust subspace estimation and clustering.
\newblock In {\em IEEE Conference on Computer Vision and Pattern Recognition},
  pages 1801 --1807, 2011.

\bibitem{Feng:CVPR14}
J.~Feng, Z.~Lin, H.~Xu, and S.~Yan.
\newblock Robust subspace segmentation with block-diagonal prior.
\newblock In {\em {IEEE} Conference on Computer Vision and Pattern
  Recognition}, pages 3818--3825, 2014.

\bibitem{Georghiades:TPAMI01}
A.-S. Georghiades, P.-N. Belhumeur, and D.-J. Kriegman.
\newblock From few to many: Illumination cone models for face recognition under
  variable lighting and pose.
\newblock {\em {IEEE} Transactions on Pattern Analysis and Machine
  Intelligence}, 23(6):643--660, 2001.

\bibitem{Guo:IJCAI15}
X.~Guo.
\newblock Robust subspace segmentation by simultaneously learning data
  representations and their affinity matrix.
\newblock In {\em Proceedings of the 24th International Joint Conference on
  Artificial Intelligence}, pages 3547--3553, 2015.

\bibitem{Hastie:StatSci98}
T.~Hastie and P.-Y. Simard.
\newblock Metrics and models for handwritten character recognition.
\newblock {\em Statistical Science}, pages 54--65, 1998.

\bibitem{He:TNNLS16}
R.~He, L.~Wang, Z.~Sun, Y.~Zhang, and B.~Li.
\newblock Information theoretic subspace clustering.
\newblock {\em IEEE Transactions on Neural Networks and Learning Systems},
  27(12):2643--2655, 2016.

\bibitem{Hinton:ISPM12}
G.~Hinton, L.~Deng, D.~Yu, G.-E. Dahl, A.-R. Mohamed, N.~Jaitly, A.~Senior,
  V.~Vanhoucke, P.~Nguyen, and T.~N. Sainath.
\newblock Deep neural networks for acoustic modeling in speech recognition: The
  shared views of four research groups.
\newblock {\em IEEE Signal Processing Magazine}, 29(6):82--97, 2012.

\bibitem{Ho:CVPR03}
J.~Ho, M.-H. Yang, J.~Lim, K.-C. Lee, and D.-J. Kriegman.
\newblock Clustering appearances of objects under varying illumination
  conditions.
\newblock In {\em Proceedings of {IEEE} International Conference on Computer
  Vision and Pattern Recognition}, pages 11--18, 2003.

\bibitem{Ji:AdaLRK-arXiv17}
P.~Ji, I.~Reid, R.~Garg, H.~Li, and M.~Salzmann.
\newblock Adaptive low-rank kernel subspace clustering.
\newblock {\em arXiv:1707.04974v4}, 2019.

\bibitem{Ji:WACV14}
P.~Ji, M.~Salzmann, and H.~Li.
\newblock Efficient dense subspace clustering.
\newblock In {\em IEEE Winter conferance on Applications of Computer Vision},
  pages 461--468, 2014.

\bibitem{Ji:NIPS17}
P.~Ji, T.~Zhang, H.~Li, M.~Salzmann, and I.~Reid.
\newblock Deep subspace clustering networks.
\newblock In {\em Neural Information Processing Systems {(NIPS)}}, 2017.

\bibitem{Krizhevsky:NIPS2012}
A.~Krizhevsky, I.~Sutskever, and G.~E. Hinton.
\newblock Imagenet classification with deep convolutional neural networks.
\newblock In {\em Neural Information Processing Systems}, pages 1097--1105,
  2012.

\bibitem{Lezama:CVPR18}
J.~Lezama, Q.~Qiu, P.~Muse, and G.~Sapiro.
\newblock Ole: Orthogonal low-rank embedding - a plug and play geometric loss
  for deep learning.
\newblock In {\em Proceedings of {IEEE} International Conference on Computer
  Vision and Pattern Recognition}, pages 8109--8118, 2018.

\bibitem{Li:CVPR15}
C.-G. Li and R.~Vidal.
\newblock Structured sparse subspace clustering: A unified optimization
  framework.
\newblock In {\em Proceedings of {IEEE} International Conference on Computer
  Vision and Pattern Recognition}, pages 277--286, 2015.

\bibitem{Li:TIP17}
C.-G. Li, C.~You, and R.~Vidal.
\newblock Structured sparse subspace clustering: A joint affinity learning and
  subspace clustering framework.
\newblock {\em {IEEE} Transactions on Image Processing}, 26(6):2988--3001,
  2017.

\bibitem{Li:JSTSP18}
C.-G. Li, C.~You, and R.~Vidal.
\newblock On geometric analysis of affine sparse subspace clustering.
\newblock {\em {IEEE} Journal on Selected Topics in Signal Processing}, 12(6),
  2018.

\bibitem{Li:ICPR18}
C.-G. Li, J.~Zhang, and J.~Guo.
\newblock Constrained sparse subspace clustering with side information.
\newblock In {\em Proceedings of the 24th International Conference on Pattern
  Recognition (ICPR)}, pages 2093--2099, August 2018.

\bibitem{Li:PR16}
Q.~Li, Z.~Sun, Z.~Lin, R.~He, and T.~Tan.
\newblock Transformation invariant subspace clustering.
\newblock {\em Pattern Recognition}, pages 142--155, 2016.

\bibitem{Liu:TPAMI13}
G.~Liu, Z.~Lin, S.-C. Yan, J.~Sun, Y.~Yu, and Y.~Ma.
\newblock Robust recovery of subspace structures by low-rank representation.
\newblock {\em {IEEE} Transactions on Pattern Analysis and Machine
  Intelligence}, 35(1):171--184, 2013.

\bibitem{Liu:ICML10}
G.~Liu, Z.~Lin, and Y.~Yu.
\newblock Robust subspace segmentation by low-rank representation.
\newblock In {\em Proceedings of International Conference on Machine Learning},
  pages 663--670, 2010.

\bibitem{Liu:ICCV11}
G.~Liu and S.~Yan.
\newblock Latent low-rank representation for subspace segmentation and feature
  extraction.
\newblock In {\em {IEEE} International Conference on Computer Vision}, pages
  1615--1622, 2011.

\bibitem{Lowe:IJCV04}
D.~Lowe.
\newblock Distinctive image features from scale-invariant keypoints.
\newblock {\em International Journal of Computer Vision}, 20:91--110, 2004.

\bibitem{Lu:ICCV13-TraceLasso}
C.~Lu, Z.~Lin, and S.~Yan.
\newblock Correlation adaptive subspace segmentation by trace lasso.
\newblock In {\em Proceedings of {IEEE} International Conference on Computer
  Vision}, pages 1345--1352, 2014.

\bibitem{Lu:ECCV12}
C.-Y. Lu, H.~Min, Z.-Q. Zhao, L.~Zhu, D.-S. Huang, and S.-C. Yan.
\newblock Robust and efficient subspace segmentation via least squares
  regression.
\newblock {\em Proceedings of European Conference on Computer Vision}, pages
  347--360, 2012.

\bibitem{McWilliams:DMKD14}
B.~McWilliams and G.~Montana.
\newblock Subspace clustering of high dimensional data: a predictive approach.
\newblock {\em Data Mining and Knowledge Discovery}, 28(3):736--772, 2014.

\bibitem{Munkres:JSIAM57}
J.~Munkres.
\newblock Algorithms for the assignment and transportation problems.
\newblock {\em Journal of the Society for Industrial and Applied Mathematics},
  5(1):32--38, 1957.

\bibitem{Nene:COIL}
S.-A. Nene, S.-K. Nayar, and H.~Murase.
\newblock Columbia object image library.
\newblock {\em Columbia University}, 1996.

\bibitem{Ngu:Neuc15}
H.~Nguyen, W.~Yang, F.~Shen, and C.~Sun.
\newblock Kernel low-rank representation for face recognition.
\newblock {\em Neurocomputing}, 155:32--42, 2015.

\bibitem{Parkhi:BMVC15}
O.~M. Parkhi, A.~Vedaldi, A.~Zisserman, et~al.
\newblock Deep face recognition.
\newblock In {\em BMVC}, volume~1, page~6, 2015.

\bibitem{Patel:JSTSP15}
V.~M. Patel, H.~V. Nguyen, and R.~Vidal.
\newblock Latent space sparse and low-rank subspace clustering.
\newblock {\em {IEEE} Journal of Selected Topics in Signal Processing},
  9(4):691--701, 2015.

\bibitem{Patel:ICCV13}
V.-M. Patel, H.~V. Nguyen, and R.~Vidal.
\newblock Latent space sparse subspace clustering.
\newblock In {\em Proceedings of {IEEE} International Conference on Computer
  Vision}, pages 225--232, Dev 2013.

\bibitem{Patel:ICIP14}
V.-M. Patel. and R.~Vidal.
\newblock Kernel sparse subspace clustering.
\newblock In {\em Proceedings of {IEEE} International Conference on Image
  Processing}, pages 2849--2853, 2014.

\bibitem{Peng:arxiv17}
X.~Peng, J.~Feng, S.~Xiao, J.~Lu, Z.~Yi, and S.~Yan.
\newblock Deep sparse subspace clustering.
\newblock {\em arXiv preprint arXiv:1709.08374}, 2017.

\bibitem{Peng:IJCAI16}
X.~Peng, S.~Xiao, J.~Feng, W.~Y. Yau, and Z.~Yi.
\newblock Deep subspace clustering with sparsity prior.
\newblock In {\em International Joint Conference on Artificial Intelligence},
  pages 1925--1931, 2016.

\bibitem{Peng:TCYB16}
X.~Peng, Z.~Yu, Z.~Yi, and H.~Tang.
\newblock Constructing the $l_2$-graph for robust subspace learning and
  subspace clustering.
\newblock {\em IEEE Transactions on Cybernetics}, 47(4):1053--1066, 2017.

\bibitem{Qi:TPAMI14}
X.~Qi, R.~Xiao, C.-G. Li, Y.~Qiao, J.~Guo, and X.~Tang.
\newblock Pairwise rotation invariant co-occurrence local binary pattern.
\newblock {\em {IEEE} Transactions on Pattern Analysis and Machine
  Intelligence}, 36(11):2199--2213, 2014.

\bibitem{Sam:94}
F.-S. Samaria and A.-C. Harter.
\newblock Harter, a.: Parameterisation of a stochastic model for human face
  identification.
\newblock In {\em Proceedings of the Second IEEE Workshop on Applications of
  Computer Vision}, pages 138--142, 1994.

\bibitem{Rao:TPAMI10}
R.~Shankar, T.~Roberto, R.~Vidal, and Y.~Ma.
\newblock Motion segmentation in the presence of outlying, incomplete, or
  corrupted trajectories.
\newblock {\em {IEEE} Transactions on Pattern Analysis and Machine
  Intelligence}, 32(10):1832--1845, 2010.

\bibitem{CT:IJCV92}
C.~Tomasi and T.~Kanade.
\newblock Shape and motion from image streams under orthography: a
  factorization method.
\newblock {\em International Journal on Computer Vision}, 9(2):137--154, 1992.

\bibitem{ReVidal:ISM11}
R.~Vidal.
\newblock Subspace clustering.
\newblock {\em IEEE Signal Processing Magazine}, 28(2):52--68, 2011.

\bibitem{ReVidal:PRL14}
R.~Vidal and P.~Favaro.
\newblock Low rank subspace clustering (lrsc).
\newblock {\em Pattern Recognition Letters}, 43:47--61, 2014.

\bibitem{Vidal:PAMI05}
R.~Vidal, Y.~Ma, and S.~Sastry.
\newblock {Generalized Principal Component Analysis (GPCA)}.
\newblock {\em {IEEE} Transactions on Pattern Analysis and Machine
  Intelligence}, 27(12):1--15, 2005.

\bibitem{Vidal:Springer16}
R.~Vidal, Y.~Ma, and S.~Sastry.
\newblock {\em Generalized Principal Component Analysis}.
\newblock Springer Verlag, 2016.

\bibitem{Wang:NIPS13-LRR+SSC}
Y.~X. Wang, H.~Xu, and C.~Leng.
\newblock Provable subspace clustering: when lrr meets ssc.
\newblock In {\em Neural Information Processing Systems {(NIPS)}}, pages
  64--72, 2013.

\bibitem{Xiao:TNNLS16}
S.~Xiao, M.~Tan, D.~Xu, and Z.-Y. Dong.
\newblock Robust kernel low-rank representation.
\newblock {\em IEEE Transactions on Neural Networks and Learning Systems},
  27(11):2268--2281, 2016.

\bibitem{Xin:TSP18}
B.~Xin, Y.~Wang, W.~Gao, and D.~Wipf.
\newblock Building invariances into sparse subspace clustering.
\newblock {\em IEEE Transactions on Signal Processing}, 66(2):449--462, 2018.

\bibitem{You:ECCV18}
C.~You, C.~Li, D.~P. Robinson, and R.~Vidal.
\newblock Scalable exemplar-based subspace clustering on class-imbalanced data.
\newblock In {\em Proceedings of European Conference on Computer Vision},
  September 2018.

\bibitem{You:CVPR16-EnSC}
C.~You, C.-G. Li, D.~Robinson, and R.~Vidal.
\newblock Oracle based active set algorithm for scalable elastic net subspace
  clustering.
\newblock In {\em Proceedings of {IEEE} International Conference on Computer
  Vision and Pattern Recognition}, pages 3928--3937, 2016.

\bibitem{You:CVPR16}
C.~You, D.~Robinson, and R.~Vidal.
\newblock Scalable sparse subspace clustering by orthogonal matching pursuit.
\newblock In {\em Proceedings of {IEEE} International Conference on Computer
  Vision and Pattern Recognition}, pages 3918--3927, 2016.

\bibitem{Zhang:VCIP16}
J.~Zhang, C.-G. Li, H.~Zhang, and J.~Guo.
\newblock Low-rank and structured sparse subspace clustering.
\newblock In {\em Proceeding of IEEE Visual Communication and Image
  Processing}, 2016.

\bibitem{Zhou:CVPR18}
P.~Zhou, Y.~Hou, and J.~Feng.
\newblock Deep adversarial subspace clustering.
\newblock In {\em Proceedings of {IEEE} International Conference on Computer
  Vision and Pattern Recognition}, June 2018.

\end{thebibliography}

\end{document}